# From Machine Learning to Machine Reasoning

Léon Bottou

2/8/2011

**Abstract**

A plausible definition of "reasoning" could be "algebraically manipulating previously acquired knowledge in order to answer a new question". This definition covers first-order logical inference or probabilistic inference. It also includes much simpler manipulations commonly used to build large learning systems. For instance, we can build an optical character recognition system by first training a character segmenter, an isolated character recognizer, and a language model, using appropriate labelled training sets. Adequately concatenating these modules and fine tuning the resulting system can be viewed as an algebraic operation in a space of models. The resulting model answers a new question, that is, converting the image of a text page into a computer readable text.

This observation suggests a conceptual continuity between algebraically rich inference systems, such as logical or probabilistic inference, and simple manipulations, such as the mere concatenation of trainable learning systems. Therefore, instead of trying to bridge the gap between machine learning systems and sophisticated "all-purpose" inference mechanisms, we can instead algebraically enrich the set of manipulations applicable to training systems, and build reasoning capabilities from the ground up.

## 1. Introduction

Since learning and reasoning are two essential abilities associated with intelligence, machine learning and machine reasoning have both received much attention during the short history of computer science. The statistical nature of learning is now well understood (e.g., Vapnik, 1995). Statistical machine learning methods are now commonplace (NIPS, 1987-2010.) An internet search for "support vector machines" returns more than two million web pages. The nature of reasoning has proven more elusive. Although computer algorithms for logical inference (Robinson, 1965) share their roots with the foundations of mathematics, converting ordinary data into a consistent set of logical expressions has proved very challenging: searching the discrete spaces of symbolic formulas often leads to a combinatorial explosion (Lighthill, 1973). Computer algorithms for general probabilistic inference (Pearl, 1988) still suffer from unfavorable computational properties (Roth, 1996). However, there are practical algorithms for many special cases of interest. These algorithms have gained considerable popularity in the machine learning community. This practicality comes at the price of reduced expressive capabilities: since probabilistic inference is a mathematical construction, it is easily described using first order logic; the converse is not true. In particular, expressing causality with first order logic is very simple, expressing causality with probabilities is challenging (Pearl, 2000).

Human reasoning displays neither the limitations of logical inference nor those of probabilistic inference.

The ability to reason is often confused with the ability to make logical inferences. When we observe a visual scene, when we hear a complex sentence, we are able to explain in formal terms the relation of the objects in the scene, or the precise meaning of the sentence components. However, there is no evidence that such a formal analysis necessarily takes place: we see a scene, we hear a sentence, and we just know what they mean. This suggests the existence of a middle layer, already a form of reasoning, but not yet formal or logical. Investigating informal reasoning is attractive because we hope to avoid the computational complexity issues associated with combinatorial searches in the vast space of discrete logic propositions.

Minsky and Papert (1969) have shown that simple cognitive tasks cannot be implemented using linear threshold functions but require multiple layers of computation. Recent advances have uncovered effective strategies to train such deep models (Hinton et al., 2006). Deep learning has attracted considerable interest in the machine learning community. Regular workshops have been held during the NIPS and ICML conferences since 2007. The ability to train deep machine learning models appears to be related to the appropriate definition of unsupervised auxiliary tasks that help discovering internal representations in the inner layers of the models (Bengio et al., 2007; Weston et al., 2008.) Deep structures had been trained in the past using supervised intermediate tasks (e.g., Bottou et al., 1997; LeCun et al., 1998.) The surprise of deep learning is that the same results can be achieved using very loosely related auxiliary tasks. Deep learning is therefore intimately related to multi-task learning (Caruana, 1997.)

This document presents an idea that results from a long maturation (Bottou, 2008). Both deep learning and multi-task learning show that we can leverage auxiliary tasks to help solving a task of interest. This apparently simple idea can be interpreted as a rudimentary form of reasoning. Enriching this algebraic structure then leads to higher forms of reasoning. This provides a path to build reasoning abilities into machine learning systems from the ground up.

## 2. Auxiliary tasks

One frequently mentioned problem is the scarcity of labeled data. This assertion is biased because we usually build a learning machine to accomplish a valuable task. The corresponding training labels are then expensive and therefore scarce. Conversely, labels available in abundance are often associated with tasks that are not very valuable. But this does not make these abundant labels useless: in the vicinity of an interesting and valuable task, less valuable tasks provide opportunities to approach the initial problem.

Consider the task of identifying persons from face images. Despite the increasing availability of collaborative image tagging schemes (von Ahn, 2006), it certainly remains expensive to collect and label millions of training images representing the face of each subject with a good variety of positions and contexts. However, it is easy to collect training data for the slightly different task of telling whether two faces in images represent the same person or not (Miller, 2006): two faces in the same picture are likely to belong to different persons; two faces in successive video frames are likely to belong to the same person. These two tasks have much in common: image analysis primitives, feature extraction, part recognizers trained on the auxiliary task can certainly help solving the original task.

Figure 1 outlines a transfer learning strategy involving three trainable modules. The preprocessor P computes a compact face representation from the image. The comparator D compares the representations associated with two images. The classifier C produces the person label associated with an image representation. We first assemble two instances of the preprocessor P and one comparator D and train the resulting model using the abundant labels for the auxiliary task. Training simply works by minimizing a regularized loss function using stochastic gradient descent. Then we assemble another instance of the preprocessor P with the classifier C and train the resulting model using a restrained number of labeled examples for the original task. This works because the preprocessor P already performs useful tasks and vastly simplifies the job of the classifier C. Alternatively we could simultaneously train both assemblages by making sure that all instances of the preprocessor share the same parameters. Comparable transfer learning systems have achieved high accuracies on vision benchmarks (e.g., Ahmed et al., 2008.)

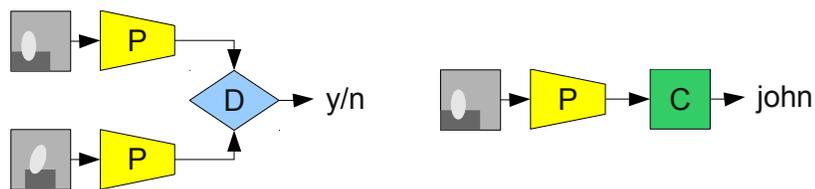

Figure 1 – Training a face recognizer.

We have designed a structurally more complex system to address various natural language processing benchmark tasks (Collobert et al., 2007, 2011). The word embedding module W computes a 50-dimensional representation for each vocabulary word. Fixed length sequences of words are extracted from a large corpus (900M words). Incorrect sequences are created by randomly replacing the central word. The auxiliary task consists in producing a score whose magnitude indicates whether a sequence of words is genuine or incorrect. An assemblage of several word embedding module W and a ranking module R is trained on this task. The benchmark tasks are then trained using smaller corpora of labelled sentences. Each sentence is processed by assembling the word embedding components W and routing their outputs, together with ancillary information, to classifiers that produce tags for the word(s) of interest (figure 2.)

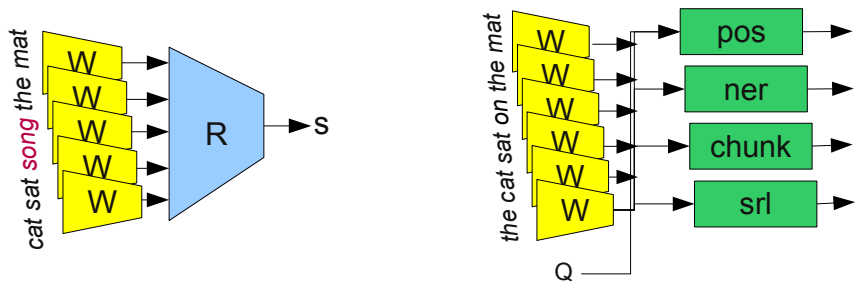

Figure 2 – Training intermediate word representations for multiple natural language processing tasks.

This system reaches near state-of-the-art performance while running hundreds of times faster than natural language processing systems with comparable performance. Many natural language processing systems rely on the considerable linguistic knowledge that went into the manual design of task specific input features. The system described above learns useful features using an essentially unsupervised task trained on a very large corpus. Figure 3 illustrates the quality of the resulting word representation.

| FRANCE | JESUS | XBOX | REDDISH | SCRATCHED | MEGABITS |
|---|---|---|---|---|---|
| 454 | 1973 | 6909 | 11724 | 29869 | 87025 |
| AUSTRIA | GOD | AMIGA | GREENISH | NAILED | OCTETS |
| BELGIUM | SATI | PLAYSTATION | BLUISH | SMASHED | MB/S |
| GERMANY | CHRIST | MSX | PINKISH | PUNCHED | BIT/S |
| ITALY | SATAN | IPOD | PURPLISH | POPPED | BAUD |
| GREECE | KALI | SEGA | BROWNISH | CRIMPED | CARATS |
| SWEDEN | INDRA | psNUMBER | GREYISH | SCRAPED | KBIT/S |
| NORWAY | VISHNU | HD | GRAYISH | SCREWED | MEGAHERTZ |
| EUROPE | ANANDA | DREAMCAST | WHITISH | SECTIONED | MEGAPIXELS |
| HUNGARY | PARVATI | GEFORCE | SILVERY | SLASHED | GBIT/S |
| SWITZERLAND | GRACE | CAPCOM | YELLOWISH | RIPPED | AMPERES |

Figure 3 – Each column shows a query word, its frequency rank in the training corpus, and the ten words whose representation is closest to that of the query word (Collobert et al., 2011.)

## 3. Reasoning revisited

Although modular learning systems and their training algorithms have been researched extensively (e.g., Bottou and Gallinari, 1991), little attention has been paid to the rules that describe how to assemble trainable modules in order to address a particular task. In fact, these composition rules play an extremely important role. The dictionary of elementary trainable modules and the composition operators form a simple algebraic system on a space of models. The composition rules describe the algebraic manipulations that let us combine previously acquired knowledge – in the form of models previously trained on auxiliary tasks – in order to create a model that addresses a new task.

I would like at this point to draw a bold parallel: "algebraic manipulation of previously acquired knowledge in order to answer a new question" is a plausible definition of the word "reasoning". There are significant differences: conventional reasoning operates on "premises and conclusions"; composition rules operate on

"trainable modules". Yet we can easily argue that the history of mathematics teaches that algebraic structures are more significant than the objects on which they operate. In both the face recognition and the natural language processing examples, the implicit composition rules derive from the assumption that internal representations that can be learned on the auxiliary task and can benefit the task of interest. These internal representations play the same role as reasoning abstractions, that is, concepts that cannot be observed directly but are assumed relevant for multiple problems.

Composition rules can be described with very different levels of sophistication. Like the face recognition and the natural language processing examples, most works discussing multi-task learning (Caruana, 1997) construct ad-hoc combinations justified by a semantic interpretation of the internal representations. Works discussing structured learning systems (e.g., Bakir et al., 2007) often provide more explicit rules. For instance, graph transformer networks (Bottou et al., 1997; LeCun et al., 1998, section IV) construct specific recognition and training models for each input image using graph transduction algorithms. The specification of the graph transducers then should be viewed as a description of the composition rules (figure 5).

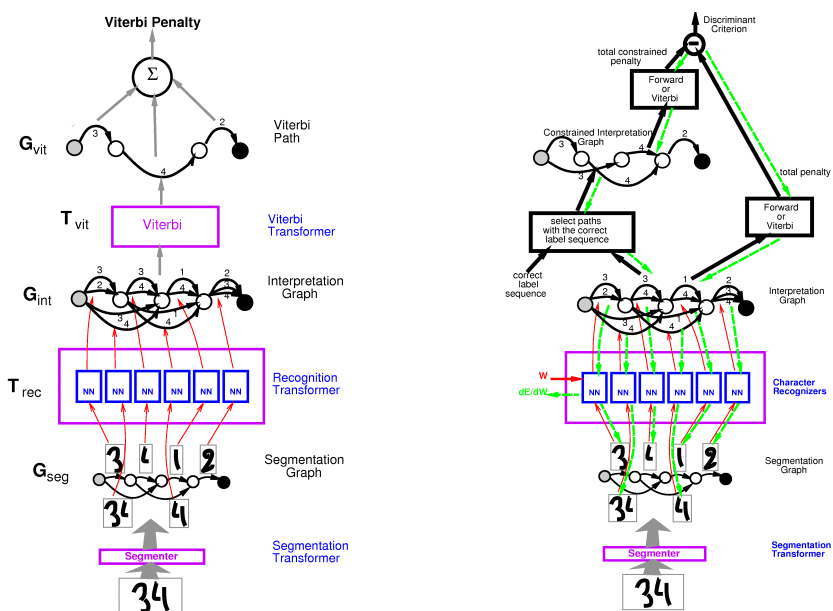

Figure 5 – Graph transformer networks for handwritten text field segmentation (LeCun et al, 1998). The drawings represent how graph transformations define the recognition and training architectures for the same example image.

## 4. Probabilistic Models

The rich algebraic structure of probability theory plays an important role in the appeal of probabilistic models in machine learning because it tells how to combine conditional probability distributions and how to interpret these combinations. However, in order to construct an algebraic structure of probabilistic models, it is necessary to also discuss how probability distributions are parametrized.

Graphical models (Pearl, 1988) describe the factorization of a joint probability distribution into elementary conditional distributions with specific conditional independence assumptions. This factorization suggests to individually parametrize these elementary conditional distribution. The probabilistic inference rules then induce an algebraic structure on the space of conditional probability distribution models describing relations between arbitrary subsets of random variables.

Many refinements have been devised to make the parametrization more explicit. The plate notation (Buntine, 1994) compactly represents large graphical models with repeated structures that usually share parameters. Figure 5 shows how treating the parameters like a random variable makes the parametrization even more explicit. More recent works propose considerably richer languages to describe large graphical probabilistic models. Probabilistic Relational Models (Friedman et al., 1999) and Relational Dependency Networks (Neville and Jensen, 2003) derive graphical probabilistic models from frame-based knowledge bases. Markov Logic Networks (Richardson and Domingos, 2006) derive graphical probabilistic models from the clauses of a first order logic knowledge base. Such high order languages for describing probabilistic models are  expressions of the composition rules described in the previous section.

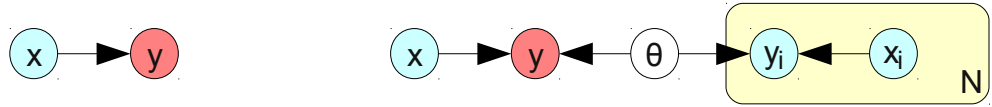

Figure 5 – Plates. The left graph simply describes the factorization $P(x,y)=P(x).P(y \mid x)$. The right graph introduces the training set $\{(x_i,y_i)\}$ and the model parameters $\theta$ using the plate notation. Bayesian inference then gives the expression of $P(y \mid x, \{(x_i,y_i)\})$.

## 5. Reasoning systems

We are clearly drifting away from the statistical approach because we are no longer fitting a simple statistical model to the data. In fact, we are dealing with a more complex object composed of (a) an algebraic space of models, and (b) composition rules that establish a homomorphic correspondence between the space of models and the space of questions of interest. We call such an object a "reasoning system."

A potentially surprising consequence of this definition is the arbitrary nature of a reasoning system. Just like statistical models, reasoning systems vary in expressive power, in predictive abilities, and in computational requirements. A few salient examples can illustrate this diversity:

- *First order logic reasoning* – Consider a space of models composed of functions that predict the truth value of a first logic formula as a function of the potential values of its free variables. Such functions are usually represented by collections of fully instantiated predicates. This space of functions is highly constrained by the algebraic structure of the first order logic formulas: if we know some of these functions, we can apply logical inference to deduct or constrain functions associated with other formulas and therefore representing different tasks. First order logic has very high expressive power because the bulk of mathematics can formalized as first order logic statements (Hilbert and Ackermann, 1928.) This is not sufficient, however, to express the subtleties of natural language: every first order logic formula is easily expressed in natural language; the converse is not true. Finally, first order logic typically leads to computationally expensive algorithms because they  often involve combinatorial searches in vast discrete spaces.

- *Probabilistic reasoning* – Consider the space of models formed by all the conditional probability distributions associated with a predefined collection of random variables. These conditional distributions are highly constrained by the algebraic properties of the probability theory: if we know a subset of these conditional distributions, we can apply Bayesian inference to deduct or constrain additional conditional distributions and therefore answer different questions (Pearl, 1988). The continuous nature of probability theory provides opportunities to avoid computationally expensive discrete combinatorial searches. These improved computational requirements come at the price of reduced expressive capabilities: since probabilistic inference is a mathematical construction, it is easily described using first order logic; the converse is not true. Despite this limitation, inference in probabilistic models is a popular machine learning topic.

- *Causal reasoning* –  Causality is a well known expressive limitation of probabilistic reasoning. For instance, we can establish a correlation between the events "it is raining" and "people carry open umbrellas". This correlation is predictive: if people carry open umbrellas, we can be pretty certain that it is raining. But this correlation tells us little about the consequences of an intervention: banning umbrellas will not stop the rain. This is a serious limitation because causes and effects play a central role in our understanding of the world. Pearl (2003) proposes to address this issue by enriching the probabilistic machinery with a new construction: whereas $P(X|Y=y)$ represents the distribution of random variable $X$ given the observation $Y=y$, the new construction $P(X|do(Y=y))$ represents the distribution of $X$ when an intervention enforces the condition $Y=y$.

- *Newtonian mechanics* – Classical mechanics is an extremely successful example of causal reasoning system. Consider the motion of point masses in various experimental setups. Newton's laws define the abstract concept of a force as the cause explaining any deviation from the uniform motion. The second and third laws then describe how to compute the consequences of interventions such as applying a new force or transferring a point mass from one experimental setup into another. For instance, a first setup could be a weighting device that measures the relative masses of point masses A and B; and a second setup could involve the collision of point masses A and B.

- *Spatial reasoning* – How would a visual scene change if one changes the viewpoint or if one manipulates one of the objects in the scene? Such questions clearly obey algebraic constraints that derive from the bi-dimensional projection and from the relations between objects. Spatial reasoning does not require the full logic apparatus but certainly benefits from the definition of specific algebraic constructions (Aiello et al., 2007.)

- *Social reasoning* – Changes of viewpoint also play an important role in social interactions. Placing oneself in somebody else's shoes allows us to understand the beliefs and the intents of other members of the society and therefore plays an essential cognitive role (Baron-Cohen, 1997). It is certainly challenging, but conceivable, to approach social interactions in terms of such processes.

- *Non-falsifiable reasoning* – History provides countless examples of reasoning systems with questionable predictive capabilities. Mythology interprets the world by applying a social reasoning systems to abstract deities. Astrology attempts to interpret social phenomena by reasoning about the motion of planets. Just like non-falsifiable statistical models, non-falsifiable reasoning systems are unlikely to have useful predictive capabilities (Popper, 1959; Vapnik, 1994.)

There are two ways to face such a universe of reasoning systems. One approach would be to identify a single reasoning framework strictly more powerful than all others. Whether such a framework exists and whether it leads to computationally feasible algorithms is unknown. Symbolic reasoning (e.g., with first order logic) did not fulfill these hopes (Lighthill, 1973). Probabilistic reasoning is more practical but considerably less expressive. The second approach is to embrace this diversity as an opportunity to better match the reasoning models to the applicative domain of interest: "when solving a given problem, try to avoid solving a more general problem as an intermediate step" (Vapnik, 1994).

It is therefore desirable to map the universe of reasoning systems. What are the potential algebraic structures? What is their footprint in terms of expressive power, suitability for specific applications, computational requirements, and predictive abilities? Unfortunately we cannot expect such theoretical advances on schedule. We can, however, nourish our intuitions by empirically exploring the capabilities of algebraic structures designed for specific applicative domains.

The replication of essential human cognitive processes such as scene analysis, language understanding, and social interactions forms an important class of applications. These processes probably include a form of reasoning because we are able, after the facts, to explain our conclusions with logical arguments. However, the actual processes usually happen without conscious involvement, suggesting that the full complexity of logic reasoning is not required. Which algebraic reasoning primitive are suitable for such applications?

The following sections describe more specific ideas investigating reasoning systems suitable for natural language processing and vision tasks. It defines trainable modules that provide the means to represent arbitrary hierarchical structures using fixed size representation vectors. The discussion includes preliminary results on natural language processing tasks and potential directions for vision tasks. Additional modules working on this space or representations are then proposed. The section concludes with the description of conceptual and algorithmic issues associated with learning algorithms operating in this space.

# 6 Association and dissociation

We have already demonstrated the possibility to learn salient word embeddings using an essentially non supervised task (Collobert et. al., 2011). Can we learn salient embeddings for any meaningful segment of a sentence?

A proven way to create a rich algebraic system is to define operations that take their inputs in a certain space and produce outputs in the same space. The number of possible concatenations and their potential depth then becomes infinite. We consider again a collection of trainable modules. The word embedding module W computes a continuous representation for each word of the dictionary. In preliminary experiments, this is a simple lookup table that specifies a vector in a 50-dimensional representation space for each word in the dictionary. The coordinates of these vectors are determined by the training algorithm. The association module is a trainable function that takes two vectors in the representation space and produces a single vector in the same space, which is expected to represent the association of the two input vectors.

Given a sentence segment composed of *n* words, figure 6 shows how *n-1* applications of the association module reduce the sentence segment to a single vector in the representation space. We would like this vector to be a representation of the meaning of the sentence. We would also like each intermediate result to represent the meaning of the corresponding sentence fragment.

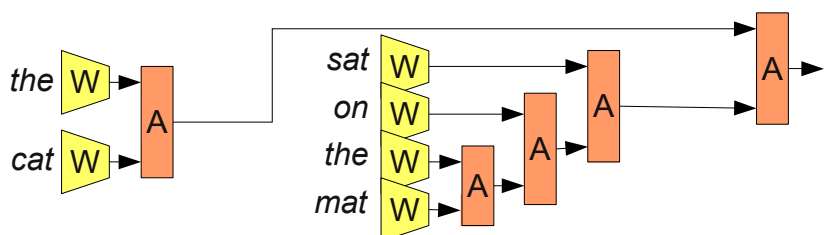

Figure 6 – Representing a sentence by recursive application of the association modules.

There are many ways to sequence the applications of the association module, each associated with a particular way of bracketing the sentence into meaningful segments. Figure 6, for instance, corresponds to the standard bracketing of the sentence "*((the cat) (sat (on (the mat))*". In order to determine which bracketing splits the sentence into the most meaningful segments, we introduce an additional saliency scoring module R, which takes as input a vector in the representation space and produces a score whose magnitude is expected to measure how meaningful is the corresponding sentence segment (figure 7).

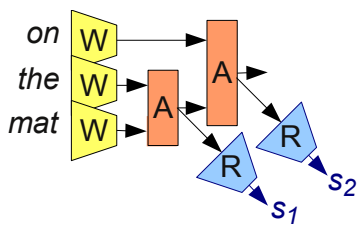

Figure 7 – The saliency scoring module R scores the quality of a sentence bracketing.

Applying the saliency scoring module R to all intermediate results and summing all the resulting scores yields a global score measuring how meaningful is a particular way to bracket a sentence. The most meaningful way to recursively apply the association module can be determined by maximizing this global score. Specific parsing algorithms are described later in this document.

The challenge is to train these modules in order to have them achieve the desired function.

Figure 8 illustrates a non-supervised training technique inspired by Collobert et. al. (2007, 2011). This is again a stochastic gradient procedure. During each iteration, a short sentence segment is randomly picked from a large text corpus and bracketed as described above (figure 8, left). An arbitrary word is then replaced by a random word from the vocabulary. As a consequence certain intermediate results in the representation space are likely to correspond to meaningless sentence fragments. We would like to make the associated scores smaller than the scores associated with the genuine sentence segments. This can be expressed by an adequate ranking loss function. The parameters of all modules are then adjusted using a simple gradient descent step. Repeating this iterative procedure corresponds to the stochastic gradient descent optimization of a well defined loss function. However, there is evidence that training works much faster if one starts with short segments and a limited vocabulary size.

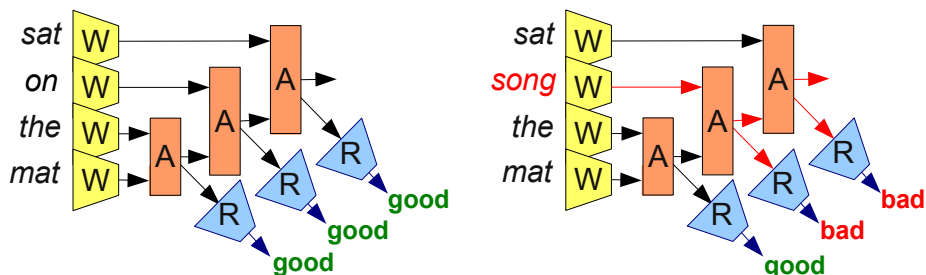

Figure 8 – Unsupervised training (see text.)
The ranking loss function tries to make the "good" scores higher than the "bad" scores.

Preliminary results have been obtained using a similar procedure (Etter, 2009). Sentence segments of length five were extracted from a dump of the English Wikipedia (600M words). The vocabulary was restricted to the 1000 most frequent words initialized with the Collobert (2011) embeddings. The initial sentence segment brackets were constructed randomly. In order to investigate how the resulting system maps word sequences into the representation space, all two-word sequences of the 500 most common words were constructed and mapped into the representation space. Figure 9 shows the closest neighbors in representation space of some of these sequences. This analysis was restricted to two-word sequences because the computational cost grows exponentially with the sequence length.

| last year   | red house    | the city    | two men        |
|-------------|--------------|-------------|----------------|
| first year  | french house | the town    | three men      |
| same year   | rock house   | the church  | four men       |
| first day   | red court    | the village | two children   |
| third year  | german house | the state   | two women      |
| first season| black house  | the country | three children |

Figure 9 – Neighbors of two-word sequences in the representation space (from Etter, 2009). Each column lists the two word sequences whose representation is closest to that of the two word sequences shown in the header.

Socher et al. (2010) independently trained a similar system in a supervised manner using the WSJ section of the annotated Penn TreeBank corpus. Although this is a much smaller corpus (about 1M words), they obtained meaningful representations. For instance they report that the phrases "*decline to comment*" and "*would not disclose the terms*" are close in the induced embedding space. The supervised training approach also provides a more objective way to assess the results since one can compare the bracketing performance of the system with that of established parsers. They report a bracketing performance that compares with that of statistical parsing systems of the 2000s.

There is much work left to accomplish, including (i) robustly addressing all the numerical aspects of the training procedure, (ii) seamlessly training using both supervised and unsupervised corpora, (iii) assessing the value of the sentence fragment representations using well known NLP benchmarks, and (iv) finding a better way to navigate these sentence fragment representations. We now introduce a new module to address this last problem.

The dissociation module D is the inverse of the association module, that is, a trainable function that computes two representation space vectors from a single vector. When its input is a meaningful output of the association module, its output should be the two inputs of the association module. Stacking one instance of the association module and one instance of the dissociation module is equivalent to an auto-encoder (figure 10, left). Recursively applying the dissociation module provides convenient means for traversing the hierarchical representations computed by a stack of association modules (figure 10, right). Such Recursive Auto-Associative Memory (RAAM) were proposed as a connectionist representation of infinite recursive structures (Pollack, 1990). Comparable ideas have been proposed by Hinton (1990).

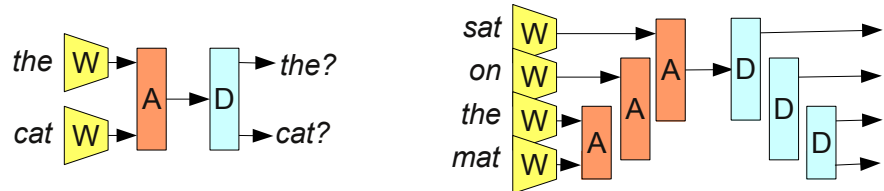

Figure 10 – Navigating intermediate representations with the dissociation module.

The domain of definition of the dissociation module is not obvious. Given a vector in representation space, we need to know whether it results from the association of two more elementary vectors or whether it should be considered as atomic. Sperduti (1994) defines and compares various labelling schemes for this purpose. However, it is probably simpler to use the saliency scoring module (figure 7) to specify the domain of definition of the dissociation module: only sufficiently meaningful associations can be dissociated.

The definition of the dissociation module implies that the association module is injective: its output uniquely defines its inputs. This property is not enforced by the training procedures outlined in the previous subsection. It is therefore necessary to simultaneously train both the association and dissociation modules. This can be achieved by augmenting the earlier loss function (figure 8) with terms that apply the dissociation module to each presumed meaningful intermediate representation and measure how close its outputs are from the inputs of the corresponding association module.

The association and dissociation modules are similar to the primitives `cons` and `car`/`cdr`, which are the elementary operations to navigate lists and trees in the Lisp computer programming languages. These primitives can be used to construct and navigate arbitrary propositional logic expressions. The main difference is the nature of the representation space. Instead of a discrete space implemented with pointers and atoms, we are using vectors in a continuous representation space. One the one hand, the depth of the structure we can construct is limited by numerical precision issues. On the other hand, numerical proximity in the representation space is meaningful (see figures 3 and 9). This property reduces the computational cost of search algorithms. This is why the multilayer stochastic gradient algorithms are able to discover meaningful intermediate representations in the first place.

Once we have constructed the means to represent arbitrary phrases using a continuous representation, we can consider training a variety of modules. Consider for instance a trainable module that converts the representation of a sentence in the present tense into a sentence in the past tense. We can then parse an initial sentence and construct its representation, convert the representation into the representation of the same sentence in the past tense, and use the dissociation module to reconstruct the converted sentence. How far we can go with such manipulations is an entirely open question.

Association and dissociation modules are not limited to natural language processing tasks. A number of state-of-the-art systems for scene categorization and object recognition use a combination of strong local features, such as SIFT or HOG features, consolidated along a pyramidal structure (e.g, Lazebnik et al., 2006). Similar pyramidal structures have long been associated with the visual cortex (Hubel and Wiesel, 1962; Riesenhuber and Poggio, 2003). Convolutional neural networks exploit the same idea (e.g., LeCun et al., 1998, 2004). Interpreting such pyramidal structures as the recursive application of an association module is relatively straightforward (e.g., Lonardi et al., 1994).

The drawback of pyramidal structures is their fixed geometry. Since local features are aggregated according to a predefined pattern, the upper levels of the pyramid represent data with poor spatial and orientation accuracy. This is why pyramidal recognition systems often work poorly as image segmentation tools. For instance, we have designed a large convolutional neural network (Grangier et al., 2009) to identify and label objects in the city scenes of the LabelMe corpus (Russel et al., 2008). This system provides good object recognition accuracies but coarse segmentations (figure 11.)

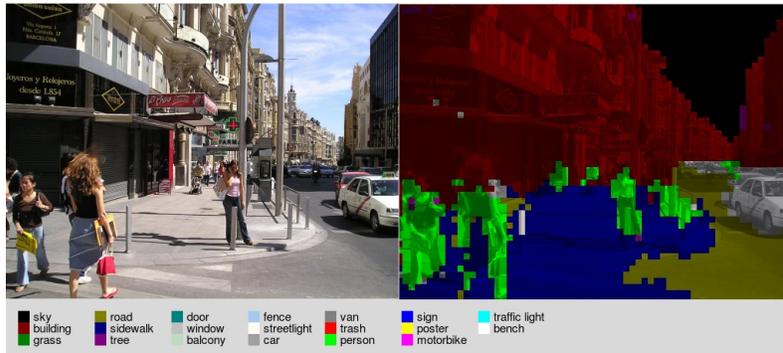

Figure 11 – Sample output of the scene labelling system described in (Grangier et al., 2009).
This large convolutional network gives good recognition accuracies but very coarse segmentations.

The parsing mechanism described for the natural language processing system provides an opportunity to work around this limitation. Let us attach intermediate representations to image regions. Initially the image regions are the small patches used for the computation of local features. Guided by a scoring module that evaluate the saliency of each potential association, the association module can then opportunistically aggregate the representations attached to two neighboring regions and produce a new representation attached with the union of the input regions. Training such a system could be achieved in both supervised and unsupervised modes, using the methods explained in the previous subsection.

Further algebraic constraints can enrich such a vision system. For instance, we could consider modules that transform vectors in representation space to account for affine transformations of the initial image. More interestingly maybe, we could consider modules that transform the representation vectors to account for changes in the position of the viewer. Since such viewpoint changes modify the occlusion patterns (e.g. Hoeim et al., 2007), such modules provides an interpretation of the three-dimensional geometry of the scene. Since viewpoint changes can also reveal or hide entire objects, such modules could conceivably provide a tool for constructing a vision system that implements object permanence (Piaget, 1937).

Finally, we can also envision modules that convert image representations into sentence representations and conversely. Training such modules would provide the means to associate sentences and images. Given an image, we could then parse the image, convert the final image representation into a sentence representation, and apply the dissociation module to reconstruct the sentence. Conversely, given a sentence, we could produce a sketch of the associated image by similar means. Much work is needed to specify the semantic nature of such conversions.

## 7. Universal parser

Let us return to the problem of determining the most meaningful way to apply the association module, which was tersely defined as the maximization of the sum of the scores computed by the ranking component for all intermediate results.

Figure 12 illustrates a maximization algorithm template whose main element is a short-term memory (STM) able to store a collection of representation vectors. The two possible actions are (1) inserting a new representation vector into the short-term memory, and (2) applying the association module A to two representation vectors taken from the short-term memory and replacing them by the combined representation vector. Each application of the association module is scored using the saliency scoring module R. The algorithm terminates when neither action is possible, that is, when the short-term memory contains a single representation vector and there are no more representation vectors to insert.

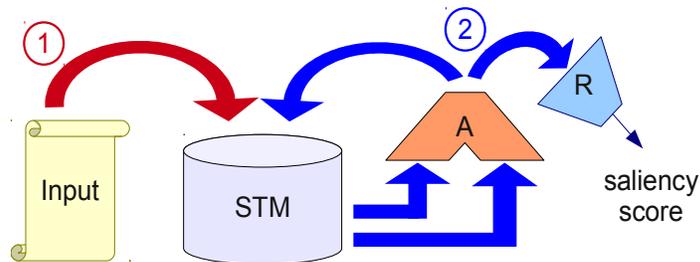

Figure 12 – The short-term memory (STM) holds a collection of representation vectors.
The two possible actions are (1) inserting a new representation vector into the STM, and
(2) replacing two vectors from the STM by the output of the association module.

The main algorithm design choices are the criteria to decide which representation vector (if any) should be inserted into the short-term memory, and which representation vectors taken from the short-term memory (if any) should be associated. These design choices then determine which data structure is most appropriate for implementing the short-term memory.

For instance, in the case of the English language, nearly all syntactically meaningful sentence segments are contiguous sequences of words. It is therefore attractive to implement the short-term memory as a stack and construct a shift/reduce parser: the first action ("shift") then consists in picking the next sentence word and pushing its representation on top of the stack; the second action ("reduce") consists in applying the association module to the top two stack elements and replacing them by the resulting representation. The problem then reduces to determining which sequence of actions to perform in order to maximize the sum of saliency scores. Even in this simple case, the graph of the possible actions grows exponentially with the length of the sentence.

Fortunately, heuristic beam search techniques are available to efficiently explore this graph. They can also handle more complicated ways to organize the short-term memory, often without dramatically increasing its computational complexity. The greedy parsing algorithm is an extreme example which consists in first inserting all word representations into the short-term memory, and repeatedly associating the two representation vectors with the highest association saliency. Simple experimentation with various compromises can suggest what works best for each application.

This parsing algorithm template is consistent with the cognitive psychology view of short-term memory. In particular, Miller (1956) argues that the human short-term memory holds seven plus-or-minus two chunks of information. Chunks are loosely defined as pieces of information that the subject recognizes as an entity. This definition depends on the knowledge of the subject, that is, the contents of her long-term memory. In the case of the parsing algorithm template, the long-term memory is represented by the trainable parameters of the association module A and the scoring module R.

## 8. More modules

The previous sections essentially discuss the association and dissociation modules. They also briefly mention a couple additional modules: modules that perform predefined transformations on natural language sentences; modules that implement specific visual reasoning primitives; and modules that bridge the representations of sentences and the representation of images. These modules enrich the algebraic reasoning structure by endowing the space of representation vectors with additional semantics.

- There is a natural framework for such enhancements in the case of natural language processing. Operator grammars (Harris, 1968) provide a mathematical description of natural languages based on transformation operators: starting from elementary sentence forms, more complex sentences are described by the successive application of sentence transformation operators. The structure and the meaning of the sentence is revealed as a side effect of these successive transformations. Since the association and dissociation modules provide the means to navigate the sentence structure, we have the necessary tools to replicate the sentence transformation operators described by Harris and establish a connection with this important body of linguistic work.

- There is also a natural framework for such enhancements in the case of vision. Modules working on representation vectors can model the consequence of various interventions. Viewpoint changes causes image rotations, image rescaling, perspective changes, and occlusion changes. We could also envision modules modeling the representation space consequences of direct interventions on the scene, such as moving an object.

There is also an opportunity to go beyond modules that merely leverage the structure of the representation space. As explained earlier, the association and dissociation modules are algebraically equivalent to the Lisp primitives `car`, `cdr`, and `cons`, and, like these primitives, provide the means to construct arbitrarily propositional logic expressions. Adding variables and quantifiers would provide an implementation of first order logic. Although there are connectionist approaches to variable binding (e.g., Smolensky, 1990), they cannot avoid the computational complexity of first-order logic problems. Would it be possible instead to identify capabilities that are necessary for the kind of informal and intuitive reasoning that humans carry out with ease? Here are two examples:

- Anaphora resolution consists in identifying which components of a tree designate the same entity. This amounts to identifying the multiple occurrences of a same variable in a first-order logic expression, or to resolving pronouns in the case of natural language sentences. This could be approached by constructing an instantiation module that takes the representation vector of a tree and applies a predefined substitution to all occurrences of a designated entity in the tree.

- Identifying which instantiations make sense could be achieved by a trainable module that returns a high score when there is a `isKindOf` relation between two representation vectors. But such ontologies are often context dependent. For instance a cat and a turtle are kinds of pets in the context of a household, and are members of different families in the context of biological classification. The restricted entailment scoring module takes the representations of two structurally similar trees and returns a high score if the first tree is a valid instantiation of the second one. This score expresses the relation between the differing tree branches in the context of the rest of the tree.

## 9. Representation space

Although the previous subsection present the essential modules as functions operating on a relatively low-dimensional vectorial space (e.g., 50-dimensional vectors), modules with similar algebraic properties could be defined on different representation spaces. Such choices have a considerable impact on the computational and practical aspects of the training algorithms. An investigation is therefore necessary.

- Our preliminary results were obtained using dense vectors with relatively low dimension, ranging from 20 to 200 dimensions (e.g., Collobert, 2011). In order to provide sufficient capabilities, the trainable functions must often be designed with a nonlinear parametrization. The training algorithms are simple extensions of the multilayer network training procedures, using gradient back-propagation and stochastic gradient descent. These nonconvex optimization procedures are inherently complex and have often been criticized for their lack of robustness. On the other hand, when properly implemented, they often turn out to be the most effective methods available for large-scale machine learning problems.

- Sparse vectors in much higher dimensional spaces are attractive because they provide the opportunity to rely more on trainable modules with linear parametrization (e.g. Paccanaro and Hinton, 2001, Mairal et al., 2010). The training algorithms can then exploit simpler optimization procedures. In order to maintain good generalization abilities and good computational performance, sparsity inducing terms must be included in the optimization criteria. Such terms also make the optimization more complex, potentially negating the benefits of sparse high-dimensional vectors in the first place.

- The representation space can also be a space of probability distributions defined on a vector of discrete random variables. The learning algorithms must then be expressed as stochastic sampling techniques such as Gibbs sampling, MCMC, Contrastive Divergence (Hinton et. al, 2006), or Herding (Welling,

2009).

Regardless of the chosen representation space, a well designed GPU implementation can considerably speed-up the experimentation cycle. For instance, training the language model of (Collobert et al., 2011) demands three to six weeks of computation on a standard processor. Reducing this training time to a couple days changes the dynamics of the experimentation.

# Conclusions

The research directions outlined in this document are intended as a breakthrough effort towards the practical and conceptual understanding of the interplay between machine learning and machine reasoning. Instead of trying to bridge the gap between machine learning systems and sophisticated "all-purpose" inference mechanisms, we can instead algebraically enrich the set of manipulations applicable to training systems, and build reasoning capabilities from the ground up. This possibility gives new ways to work around the limitations of both logical and probabilistic inference. Is this new path to Artificial Intelligence?

# Acknowledgments

I would like to acknowledge nearly three years of discussions with Yoshua Bengio, Yann LeCun and my former NEC Labs colleagues Ronan Collobert, David Grangier and Jason Weston.